\pdfoutput=1

\documentclass[11pt]{article}

\usepackage[final]{acl}

\usepackage{times}
\usepackage{latexsym}
\usepackage[T1]{fontenc}

\usepackage[utf8]{inputenc}
\usepackage{lipsum}

\newcommand\blfootnote[1]{%
  \begingroup
  \renewcommand\thefootnote{}\footnote{#1}%
  \addtocounter{footnote}{-1}%
  \endgroup
}

\usepackage{microtype}

\usepackage{inconsolata}
\usepackage{multirow} 
\usepackage{multicol} 
\usepackage{hyperref}
\usepackage{cleveref}
\usepackage{soul}
\usepackage{graphicx} 
\usepackage{amssymb} 
\title{SemEval-2024 Task 9: BRAINTEASER: A Novel Task Defying Common Sense}

\author{Yifan Jiang$^1$,
Filip Ilievski$^{1,2}$,
Kaixin Ma$^{3}$ \\
$^1$
\normalsize{Information Sciences Institute, Viterbi School of Engineering, University of Southern California}\\
$^2$\normalsize{Department of Computer Science, Faculty of Science, Vrije Universiteit Amsterdam}\\
$^3$\normalsize{Tencent AI Lab, Bellevue, WA}\\
  \texttt{yifjia@isi.edu, f.ilievski@vu.nl},
  \texttt{ kaixinma@global.tencent.com}}

\begin{document}
\maketitle
\begin{abstract}
While vertical thinking relies on logical and commonsense reasoning, lateral thinking requires systems to defy commonsense associations and overwrite them through unconventional thinking. Lateral thinking has been shown to be challenging for current models but has received little attention. A recent benchmark, BRAINTEASER, aims to evaluate current models' lateral thinking ability in a zero-shot setting.  In this paper, we split the original benchmark to also support fine-tuning setting and present SemEval Task 9: \textbf{BRAINTEASER(S)},\footnote{We use BRAINTEASER to represent the original benchmark and BRAINTEASER(S) to represent the data in SemEval task for clarity.} the first task at this competition designed to test the system's reasoning and lateral thinking ability. As a popular task, BRAINTEASER(S)'s two subtasks receive 483 team submissions from 182 participants during the competition. This paper provides a fine-grained system analysis of the competition results, together with a reflection on what this means for the ability of the systems to reason laterally. We hope that the BRAINTEASER(S) subtasks and findings in this paper can stimulate future work on lateral thinking and robust reasoning by computational models.
\end{abstract}

\section{Introduction}
\begin{figure}[!t]
	\includegraphics[width=0.47\textwidth]{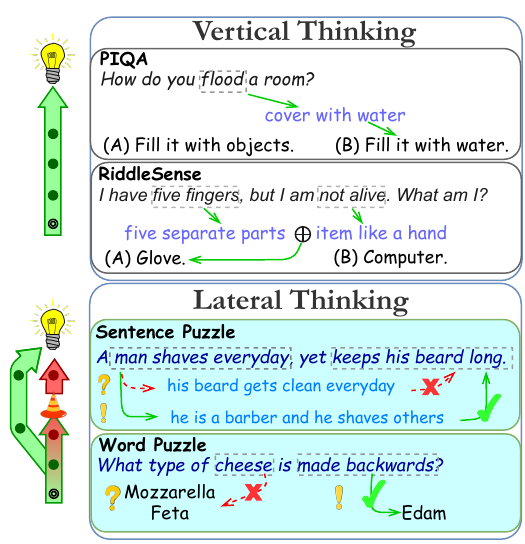}
	\caption{Figure from the first lateral thinking benchmark BRAINTEASER~\cite{jiang-etal-2023-brainteaser}, contrasting existing Vertical Thinking tasks (PIQA~\cite{bisk2020piqa} and RiddleSense~\cite{lin2021riddlesense}) to lateral thinking. Solving BRAINTEASER's lateral puzzles requires default commonsense thinking to be deprecated.}  
	\label{fig:example}
\end{figure}

Vertical thinking requires logical and commonsense reasoning, i.e., making plausible sequential associations of different pieces of commonsense knowledge. 
As presented in \autoref{fig:example} (top), we can easily infer that flooding a room requires filling it with water, based on common sense, and inanimate objects with five fingers are gloves in the riddle. 
In contrast, lateral thinking is a creative and divergent process that requires thinking out of the box and defying common sense. 
For example, as shown in~\autoref{fig:example} (bottom), one needs to overwrite the commonsense associations of \textit{man shaves} to \textit{he shaves himself}, and regard the man as somebody who shaves others all day (e.g., a barber) to answer the lateral puzzle.


While there are many datasets focusing on commonsense reasoning \cite{talmor2019commonsenseqa,bisk2020piqa,sap2019social} and numerous studies on improving commonsense reasoning ability of artificial systems \cite{ma2021knowledge,ma2021exploring,zhangstudy}, lateral thinking challenges have received little attention and are often filtered out as noise during preprocessing~\cite{vajjala2012improving,speer2017conceptnet,Sap2019ATOMICAA}. Consequently, artificial systems' ability to solve lateral thinking problems remains understudied. 


To bridge this gap, in~\cite{jiang-etal-2023-brainteaser}, we introduce a novel \textsc{BrainTeaser} benchmark with two tasks of different granularity: Sentence Puzzles and Word Puzzles (cf. \autoref{fig:example}). 
The task is formulated in a multiple-choice QA setting for a straightforward human and automatic evaluation. The dataset is constructed via a three-stage pipeline to ensure that the questions are valid and challenging. 

We organize our SemEval Task with \textbf{BRAINTEASER(S)}, which contains the same data as the BRAINTEASER 
benchmark to \textit{study model's lateral thinking ability}. Differing from the original benchmark that only focuses on the zero-shot setting, BRAINTEASER(S) divides this data into train/trial/test sets and has no limitation on the method adaptation. The goal of this paper is to describe the SemEval task and provide an analysis of the participant results. We provide details of the data construction pipeline in Section~\ref{data construction} and the SemEval Task description in Section~\ref{task description}. We present the overall leaderboard result and fine-grained method analysis in Section~\ref{result}. Finally, we discuss the summarized result and conclude with high-level insight to stimulate future works on lateral thinking. For further information, we refer the reader to our source code,\footnote{https://github.com/1171-jpg/BrainTeaser} task website,\footnote{https://brainteasersem.github.io/} and 
competition website.\footnote{https://codalab.lisn.upsaclay.fr/competitions/15566}

\section{Source Dataset}\label{data construction}
We use our recently introduced BRAINTEASER dataset \cite{jiang-etal-2023-brainteaser} as the basis for our evaluation. In this section, we briefly describe the data construction pipeline and we refer interested readers to ~\cite{jiang-etal-2023-brainteaser} for full details. 

The data construction pipeline has three stages. In the first stage, we collect lateral thinking puzzles from public websites such as \url{riddles.com} and \url{rd.com} and conduct filtering and deduplication. Then, the remaining questions are manually verified to ensure that they fit in the sentence or word puzzle categories. 

Since the collected puzzles are open-ended questions, which poses great challenges for evaluation. These open-ended puzzles are then converted to multiple-choice questions in the second stage. Specifically, we leverage tools such as COMET \cite{Hwang2021COMETATOMIC2O}, WordNet and Wikipedia to construct distractors for every question. For sentence puzzles, we collect distractors that overwrite non-central premises of the question, and for word puzzles, we collect distractors that are semantically similar to the correct answer to ensure they are challenging for systems. 

Finally, in stage three, we construct additional data to mitigate the risk of memorization by large pretrained language models. In particular, for each question, we rephrase the original question using an open-source rephrasing tool without changing its answers or distractors.\footnote{\url{https://quillbot.com/}} This set is referred to as \textit{Semantic Reconstruction}. Additionally, we leverage GPT-4 to reconstruct each question into a new context such that the misleading question premise is kept. In this case, both the question and the correct answer become different, but the reasoning path remains the same. After reconstruction, the distractors are collected in the same way as described earlier. This set is referred to as \textit{Context Reconstruction}. A strong reasoning model is expected to solve all variants of the question consistently, as their reasoning patterns are identical despite being phrased differently. In total, we construct 1,119 data samples, including reconstruction variants. We report the key statistics in~\autoref{tab:statistic}.

\begin{table}
  \begin{center}
  \small
    \caption{Key statistics of the BRAINTEASER dataset. Choices combine the correct answer with all the distractors.
    }
    \label{tab:statistic}
    \begin{tabular}{|l|c|c|} 
    \hline
       & \textbf{Sentence} & \textbf{Word}  \\  
    \hline
    \# Puzzles & 627 & 492  \\ 
    Average Question Tokens & 34.88 & 10.65 \\
    \% Long Question (>30 tokens)& 48.32\% & 2.23\%\\
    Average Answer Tokens & 9.11 & 3.0 \\ 
    Std of Choice Tokens &2.36&0.52\\
    \hline
    \end{tabular}
  \end{center}
\end{table}

\section{Task Description}\label{task description}
\subsection{Task Definition and Organization}
In BRAINTEASER(S), we utilize both subtasks in the BRAINTEASER benchmark for evaluation: Sentenze Puzzle~(\textit{SP}) and Word Puzzle~(\textit{WP}). Both subtasks are multiple-choice QA tasks. We run our SemEval task on CodaLab. Our task is divided into two primary phases: (i) The Practice Phase runs from September 2023 to January 2024, and (ii) The Evaluation Phase runs from 10th Jan 2024 to 31st Jan 2024. We open the Post-Evaluation Phase after 31st Jan 2024 to encourage further research. 
\subsection{Evaluation Metrics and Data Splits}
\noindent \textbf{Evaluation Metrics} We evaluate all systems using the same accuracy metrics as~\citet{jiang-etal-2023-brainteaser}: \textit{Instance-based Accuracy} considers each (original or reconstruction) question separately. We report instance-based accuracy on the original puzzles and their semantic and context reconstructions. \textit{Group-based Accuracy} considers each original puzzle and its variants as a group. The model will score 1 only when it successfully solves all three puzzles in the group, otherwise, its score is 0. \textit{Overall Accuracy} computes accuracy over all instances. \\
\noindent \textbf{Data Split} To enable BRAINTEASER(S) to support both fine-tuning and zero/few-shot setting, we further divided the original BRAINTEASER dataset into 3 data splits: train, trial, and test set, as shown in~\autoref{tab:data_statistic}. The train set consists of 507 sentence puzzles and 396 word puzzles. We reuse a portion of the train set as a trial set, which contains 120 sentence puzzles and 96 word puzzles. The test set has 120 data for both subtasks. We release questions and answers from the train and trial set during the Practice Phase. We only release the questions of the test set during the Evaluation Phase and release the whole dataset after the Evaluation Phase ends. \\
\noindent \textbf{Baseline} We provide three baselines~(\autoref{tab:data_statistic}, see \autoref{sec:Codalab} for details) to show the gap between humans and SOTA models. To get a comprehensive and robust evaluation performance for each subtask, the human evaluation is computed over 102
data randomly sampled from the original \textbf{BRAINTEASER} benchmark, ChatGPT and RoBERTa-L~\cite{liu2019roberta} performance are also computed over the \textbf{BRAINTEASER} in zero-shot setting, i.e. the original unpartitioned data of \cite{jiang-etal-2023-brainteaser}.

\begin{table}
\caption{Data statistics of each data split and baseline of BRAINTEASER(S).} 
\label{tab:data_statistic}
\small
\begin{center}
\begin{tabular}{l|c|c}
  & SP & WP  \\
 \hline
   BRAINTEASER& 627 & 492  \\
  \hline
 \multicolumn{3}{l}{Data Split of BRAINTEASER(S)} \\
 \hline
 Train & 507  & 396 \\
 $\hookrightarrow$ Trial~(\textit{subset of train}) &  120 & 96 \\
 Test &  120 & 120  \\
  \hline
 \multicolumn{3}{l}{Baseline overall accuracy} \\
 \hline
 Human & 0.920  &  0.917\\
 ChatGPT~(BRAINTEASER) &  0.627& 0.535\\
 RoBERTa-L~(BRAINTEASER) &  0.434& 0.207\\
 \hline
\end{tabular}
\end{center}
\end{table}

\section{Participant System and Results}\label{result}
\begin{table*}[h]
  \setlength\tabcolsep{5pt}
    \caption{Top ten leaderboard results for both subtasks, including user submissions without system description papers. Ori = Original, Sem = Semantic, Con = Context. Team name with (*) submit the system description paper. The first, second and third submissions per category are represented by \hl{highlight}, \textbf{bold} and  \underline{underline}, respectively.} 
   \label{tab:top_table}
  \begin{center}
   \small
   \begin{tabular}{ l |c | ccc|cc}
   \hline
    \multirow{2}{*}{\textbf{Team Name}} & \multirow{2}{*}{\textbf{Overall}} & \multicolumn{3}{c|}{Instance-based} & \multicolumn{2}{c}{Group-based} \\
     \cline{3-7}
  &  & \textbf{Original}     &    \textbf{Semantic}   &    \textbf{Context}  &  \textbf{Ori \& Sem }  &  \textbf{Ori \& Sem \& Con }           \\              
   \hline
      \multicolumn{7}{c}{\textbf{\textit{Sentense Puzzle}}} \\
      \hline
        abdelhak*& \hl{0.983} & \hl{1.000} & \hl{1.000} & \hl{0.950} & \hl{1.000} & \hl{0.950} \\
        HW-TSC*& \textbf{0.967} & \hl{1.000} & \textbf{0.975} & \textbf{0.925} & \textbf{0.975} & \underline{0.900}  \\
        Maxine& \underline{0.958} & \textbf{0.975} & \textbf{0.975} & \textbf{0.925} & \underline{0.950} & \underline{0.900}  \\
        YingluLi& 0.950 & \textbf{0.975} & \underline{0.950} & \textbf{0.925} & \underline{0.950} & \underline{0.900}  \\
        Theo& 0.950 & \underline{0.950} & \underline{0.950} & \hl{0.950} & \underline{0.950} & \textbf{0.925}  \\
        somethingx95& 0.942 & \underline{0.950} & \underline{0.950} & \textbf{0.925} & \underline{0.950} & \underline{0.900}  \\
        gerald& 0.942 & \underline{0.950} & \underline{0.950} & \textbf{0.925} & \underline{0.950} & \underline{0.900}  \\
        AmazUtah\_NLP*& 0.925 & 0.925 & \underline{0.950} & \underline{0.900} & 0.925 & 0.875  \\
        BITS\,Pilani*& 0.900 & \textbf{0.975} & 0.925 & 0.800 & 0.925 & 0.775  \\
        ALF*& 0.900 & 0.925 & \underline{0.950} & 0.825 & 0.925 & 0.825  \\
    \hline
    \multicolumn{7}{c}{\textbf{\textit{Word Puzzle}}} \\
      \hline
        Theo& \hl{0.990} & \hl{1.000} &\hl{1.000} & \textbf{0.969} & \hl{1.000} & \hl{0.969} \\
        gerald& \hl{0.990} & \hl{1.000} & \hl{1.000} & \textbf{0.969} & \hl{1.000} & \hl{0.969}   \\
        somethingx95& \textbf{0.979} & \hl{1.000} & \hl{1.000} & \underline{0.938} & \hl{1.000} & \textbf{0.938}  \\
        zero\_shot\_is\_all\_you\_need*& \textbf{0.979} & \hl{1.000} & \hl{1.000} & \underline{0.938} & \hl{1.000} & \textbf{0.938}  \\
        MasonTigers*& \textbf{0.979} & \textbf{0.969} & \textbf{0.969} & \hl{1.000} & \textbf{0.969} & \hl{0.969}   \\
        HW-TSC*& \underline{0.969} & \textbf{0.969} & \underline{0.938} & \hl{1.000} & \underline{0.938} & \textbf{0.938}  \\
        Maxine& \underline{0.969} & \textbf{0.969} &\underline{0.938} & \hl{1.000} & \underline{0.938} & \textbf{0.938}  \\
        YingluLi& \underline{0.969} & \textbf{0.969} & \underline{0.938} & \hl{1.000} & \underline{0.938} & \textbf{0.938}  \\
        kubapok& 0.948 & 0.906 & \hl{1.000} & \underline{0.938} & 0.906 & \underline{0.844}  \\
        BITS Pilani*& 0.917 & \underline{0.938} & \underline{0.938} & 0.875 & \underline{0.938} & 0.812  \\
    \hline 

    \end{tabular}
  \end{center}
\end{table*}
\subsection{Participant Overview}
We have 182 participants in total. In the Practice Phase, we have no limitation on the number of submissions to support exploration and enable participants to understand the submission format. We receive 243 submissions for \textit{SP} and 155 for \textit{WP}. In the Evaluation Phase, we allow up to three submissions per team and keep the submission with the best overall accuracy. Our final leaderboard has 48 team submissions for \textit{SP} and 37 for \textit{WP}. 
\subsection{Leaderboard Results}
\label{leaderboard}
\autoref{tab:top_table} (see~\autoref{sec:Codalab} for full table) displays the top ten models for each subtask, ranked by overall accuracy.  The best-performing model in \textit{SP} excels in all six metrics, whereas the leading models in \textit{WP} excel in all but context reconstruction. In the \textbf{instance-based accuracy metrics}, most top-performing models~(75\%) in two subtasks show better performance on original and semantic reconstruction compared to context reconstruction. Most models~(80\% in \textit{SP}; 70\% in \textit{WP}) show the same trend across the entire leaderboard. In the \textbf{group-based accuracy metric}, half of the top models in both tasks align with their original instance-based accuracy for the grouped original and semantic reconstruction (Ori\&Sem). Only one model in \textit{WP} maintains its performance on all reconstructions (Ori\&Sem\&Con). Across the leaderboard, more than 80 percent of models in both subtasks show a decrease in Ori\&Sem accuracy, ranging from 0.025 to 0.175 in \textit{SP} and 0.031 to 0.281 in \textit{WP}. Nearly all models show a significant drop in Ori\&Sem\&Con accuracy, with declines varying from 0.025 to 0.275 in \textit{SP} and 0.031 to 0.344 in \textit{WP}.

\subsection{Fine-grained System Analysis}

\begin{figure}[t!]
	\includegraphics[width=0.47\textwidth]{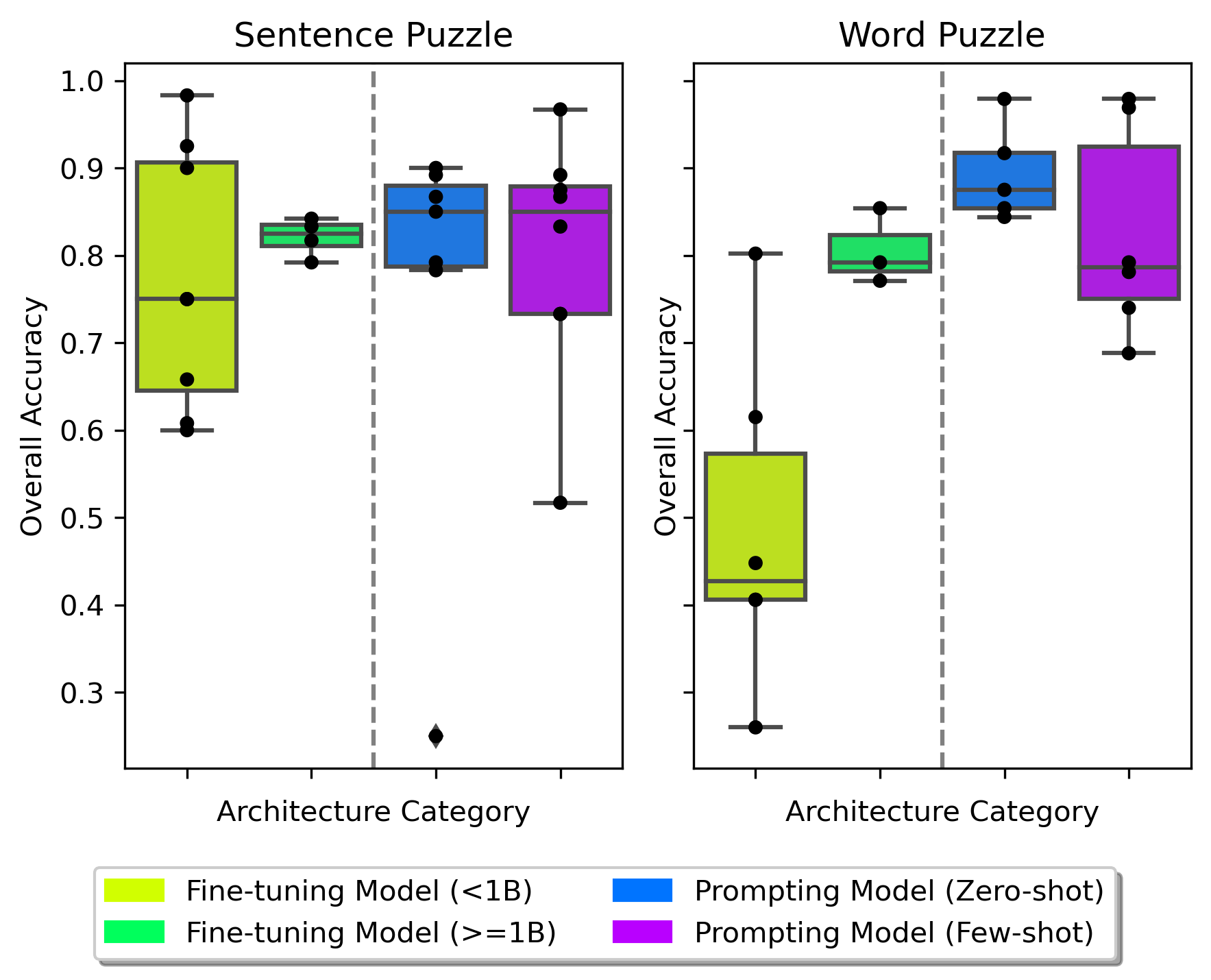}
	\caption{The overall accuracy distribution of each  architecture selection.} 
	\label{fig:architecture}
\end{figure}

In this section, we provide system analysis for the models from the 28 system description papers from participants.*\blfootnote{* The rank discussed later in this section is based on systems with description papers. }\\
\noindent \textbf{Method Adaptation and Architecture Selection} \label{architecture}
For both subtasks, the chosen adaptation methods among participants are either fine-tuning models~(60\%) or prompting models~(65\%) in a zero-shot~\cite{sanh2021multitask} or few-shot manner~\cite{brown2020language}. Half of the participants try multiple adaptations and submit the best one. For the fine-tuning architecture, participants select either small-size models~(<1B) including BERT~\cite{devlin2018bert}, RoBERTa~\cite{liu2019roberta}, DeBERTa~\cite{he2020deberta} or large-size models~(>=1B) such as FLAN-T5~\cite{chung2022scaling} and Mistral 7B~\cite{jiang2023mistral}. For the prompting architecture, the majority~(90\%) use closed-source LLMs such as GPT-4~\cite{openai2023gpt4}, GPT-3.5, GeminiPro~\cite{geminiteam2023gemini}, Claude~\cite{Claude}, and Copilot.\footnote{https://copilot.microsoft.com/} Techniques like Chain-of-Thought~\cite{wei2022chainofthought}, Ensemble~\cite{wang2022self}, and RECONCILE~\cite{chen2023reconcile} are widely adopted for prompt engineering. Figure~\ref{fig:architecture} provides a visualization of the overall accuracy distribution for each architecture. For fine-tuning architecture, fine-tuning on large models shows better performance with a tight accuracy range compared to small ones. Fine-tuning on small models shows competitive performance~(three in the top five*) in \textit{SP} but a significant drop in \textit{WP}. Among the prompting designs, both zero-shot and few-shot show promising results~(seven in the top nine systems*) on two subtasks, with the latter one having a wider accuracy range.


\noindent \textbf{External Dataset}
Half of the participants~(54\%) implement their systems only on the original target task, but some further introduce external datasets~(35\%) to enhance their models' performance. Participants generate humor-style synthetic data using LLMs, crawl riddle websites, or use RiddleSense~\cite{lin2021riddlesense} to invoke models' lateral thinking abilities.  Other commonsense datasets such as BIRD-QA~\cite{chen2021bird} or knowledge graphs including ConceptNet~\cite{speer2017conceptnet} and WordNet~\cite{miller1995wordnet} are used to provide general concepts of key instances in questions. Using humor-style datasets tends to be useful on both subtasks, especially for fine-tuning models. Meanwhile, synthetic explanations derived from LLMs are used in prompting to evoke chain-of-thought~\cite{wei2022chain} reasoning abilities.

\noindent \textbf{Data Reconstruction}
Some participants~(18\%) reconstruct the original data or change the four-choice question format. \citeauthor{wang-wang-zhang:2024:SemEval2024}~(\citeyear{wang-wang-zhang:2024:SemEval2024}) use back translation to enlarge the dataset size. \citeauthor{chakraborty-rahman-faruqe:2024:SemEval2024}~(\citeyear{chakraborty-rahman-faruqe:2024:SemEval2024}) simplify each question into the binary choice problem and \citeauthor{reyes-ramosflores-martnezmaqueda:2024:SemEval2024}~(\citeyear{reyes-ramosflores-martnezmaqueda:2024:SemEval2024}) solve the question under a classification approach with three class labels. Removing the unsure choice is also widely adopted for prompting, where the systems only choose unsure when they fail on the other three choices. Due to a limited number of data reconstruction samples, we cannot conclude which approach can improve performance.

\begin{figure}[!t]
    \centering
	\includegraphics[width=0.48\textwidth]{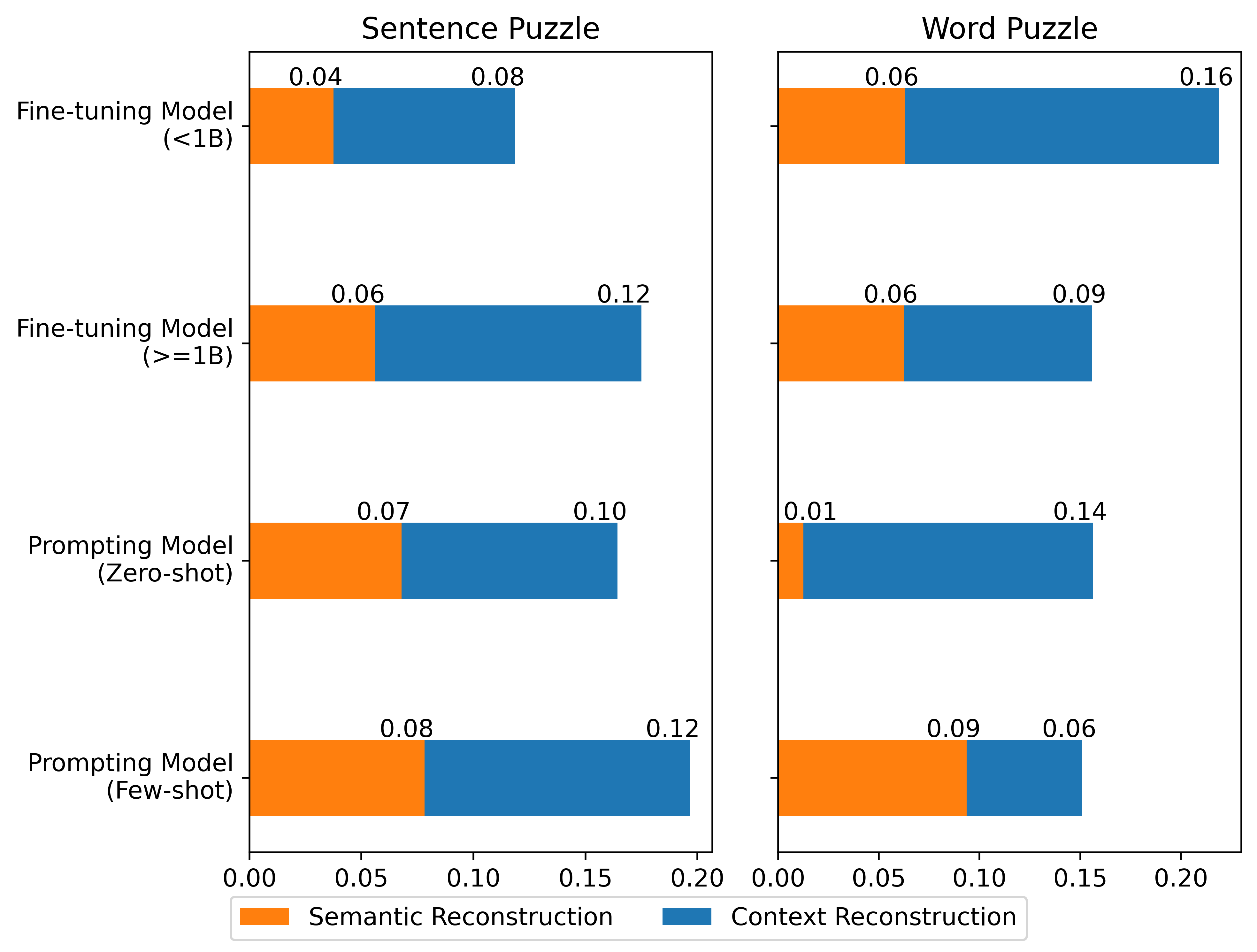}
	\caption{The drop in performance after introducing each reconstruction in group metric.} 
	\label{fig:consistency}
\end{figure}

\noindent \textbf{Consistency of Model Predictions}
In~\autoref{fig:consistency}, we compare the drop in performance when considering reconstruction variants with group metrics to understand whether the models can solve lateral thinking puzzles by following a
consistent reasoning path. On semantic reconstructions, the fine-tuning model has a smaller drop than zero/few-shot prompting in general. Fine-tuning on small models and zero-shot prompting work best on each subtask. On context reconstruction, all architectures show a more significant decline in performance. Fine-tuning on small models and few-shot prompting yield minimal drops in \textit{SP} and \textit{WP}, yet exhibit the largest declines in other subtasks.
\section{Discussion}\label{discussion}
We start the discussion with the question: ``\textit{Is lateral thinking solved}?'' The best-performing systems reach 100\% on both tasks, making it seem that the task is solved. However, there remain many questions to explore.
Our discussion targets 5 questions to provide overall insights:~1) What's the difference between the \textbf{BRAINTEASER(S)}~SemEval Task and the original \textbf{BRAINTEASER} benchmark?~2) What's the difference between the best systems for sentence puzzles and word puzzles?~3) Are model predictions consistent with individual and group partitions?~4) What does fine-tuning mean for lateral thinking tasks?~5) What challenges still exist in the realm of lateral thinking?


\subsection{Difference with the Original BRAINTEASER~\cite{jiang-etal-2023-brainteaser}} 
\label{sub:difference with bt}
The \textbf{BRAINTEASER} benchmark~\cite{jiang-etal-2023-brainteaser} is proposed to evaluate LLMs' lateral thinking ability in \textbf{zero- and few-shot} settings while in \textbf{BRAINTEASER(S)} we release 80 percent of the data for training and we put no limitation on method adaptation. Although releasing data encourages more possibilities for participants, it also narrows down our hidden test set, making the comparison between system performance on BRAINTEASER(S) and the LLMs evaluation results on the BRAINTEASER benchmark unfair. 
With only 120 samples in the BRAINTEASER(S) test set, the probability of achieving high performance by some of the large number of systems becomes relatively large. Moreover, we expect that most of the lateral patterns will be recurring between the training and the test data, which especially benefits fine-tuning methods. 
With these caveats in mind, we hope the result and analysis on BRAINTEASER(S) can provide meaningful ideas and insight on lateral thinking and be verified systematically on the whole BRAINTEASER benchmark.

\subsection{Effective System Choices and Differences}
\label{sub: effective system}
From~\autoref{architecture}, we know architecture selection yields different distributions of performances on each subtask. On sentence puzzles, fine-tuning small models~\cite{kelious-okirim:2024:SemEval2024,mishra-ghashami:2024:SemEval2024,farokh-zeinali:2024:SemEval2024} with additional dataset providing competitive results. On word puzzles, either zero-shot~\cite{moosavimonazzah-feghhi:2024:SemEval2024,venkatesh-sharma:2024:SemEval2024} or few-shot~\cite{li-EtAl:2024:SemEval20245,raihan-EtAl:2024:SemEval2024} prompting leads to top-performing results. In general, even small models obtaining language understanding during pre-training can adapt to sentence puzzles via fine-tuning, and additional humor-style datasets can evoke more lateral thinking abilities. On word puzzles, fine-tuned models have difficulties focusing on letter composition which hugely deviates from their pertaining dataset. Even the top-scoring fine-tuning model~\cite{kelious-okirim:2024:SemEval2024} on \textit{SP} fails to perform well on \textit{WP}. On the other hand, the prompting method leverages the information stored in LLMs' parameters and their access to large pre-training data to mitigate the difficulty of word puzzles. However, the nature of the frozen model not only reduces the effectiveness of the external datasets but also limits further improvement and requires meticulous prompting engineering to ensure stable performance.

\subsection{Prediction Consistency}
\label{sub: consistency}
Reconstruction of the original brainteaser puzzles allows us to distinguish between memorizing the training corpus and the ability of models to generalize to unseen samples. As indicated in~\autoref{leaderboard}, most models struggle with consistent lateral thinking. Context reconstruction poses greater challenges than semantic reconstruction due to the need for lateral reasoning adaptation to novel settings. Context reconstruction of word puzzles is the most challenging, highlighting the risks of over-fitting and memorization.~\autoref{fig:consistency} shows architectures have different consistency issues. Fine-tuned models have a significant drop in context reconstruction in \textit{WP} because the novelty of puzzles limits models to training corpus. Few-shot prompting can be beneficial for consistency in word puzzles but useless in sentence puzzles. LLMs' ability to follow pattern~\cite{mirchandani2023large} leads them to focus on the surface form in word puzzles, which brings improvement in consistency. Few-shot prompting can hardly provide general patterns of sentence puzzles due to its uniqueness, and the example in the demonstration can mislead the model.

\subsection{Impact of Fine-Tuning}
\label{sub: fine-tuning}
Even though recently in-context learning~(ICL)~\cite{brown2020language} has achieved great progress on reasoning tasks~\cite{talmor2019commonsenseqa,bisk2020piqa}, we are happy to see half of the participants implement their system in fine-tuning approaches and showing promising performance. Fine-tuning on small models can lead to a wide accuracy distribution, which requires careful design on hyperparameters and the training process. Exposure to external datasets can stabilize and enhance performance. Fine-tuning on large models shows tight accuracy distribution but lacks top-performing models, which suggests the need for more fine-tuning data to ``distort'' the default commonsense~\cite{kumar2022fine} and evoke lateral thinking out-of-distribution~\cite{jiang2023transferring}. Also, the large gap between instance- and group-based metric~(\autoref{fig:consistency}) points out that short-cut learning still exists among these methods.

\subsection{Challenges in Lateral Thinking}
We summarize the discussion with the challenges that remain unsolved and require further effort to evoke the models' lateral thinking abilities. 1)~The system performances and our analysis are based on a small set of original BRAINTEASER benchmark~(\autoref{sub:difference with bt}). A more general and systematic analysis should be performed with the entire original BRAINTEASER data or even an enlarged version of it, starting from prompting models. 2)~There is still a lack of a general approach demonstrating a stable and competitive performance on both subtasks. No existing method can merge the advantages of each architecture on each subtask~(\autoref{sub: effective system}). 3)~Each model fails to generate consistent predictions similar to humans, even under simple semantic reconstructions~(\autoref{sub: consistency}). 4)~Fine-tuning methods suffer from learning shortcuts while prompting methods have problems finding general lateral thinking patterns akin to humans (see also~\cite{lewis2024using})~(\autoref{sub: fine-tuning}).

\section{Conclusions and Future Perspectives}
This paper summarizes SemEval 2024 Task 9, BRAINTEASER(S), a novel task defying common sense.
We present the motivation, data design, data construction, evaluation process, competition systems, participant results, result analysis, and discussion.
BRAINTEASER(S) was popular among participants and received 483 submissions from 182 teams during the competition, with various method adaptations and architecture selections demonstrating different advantages on each subtask and evaluation metric. The best-performing systems have impressive performance on both sub-tasks, which reach 100\% accuracy on lateral thinking puzzles from the web. However, our fine-grained analysis highlights the remaining questions and challenges for further research. Importantly, BRAINTESER(S) SemEval result is evaluated over a subset~(20\%) of original \textsc{BrainTeaser} benchmark. Even on this subset and despite the access to 80\% of the data for training, models still struggle to reason consistently on semantic and context reconstruction.
Future work should investigate flexible ways to combine lateral and vertical thinking, construct better evaluation metrics for creative and open-ended generations, build connections within reconstruction based on analogical reasoning~\cite{sourati2023arn} and explore a dynamic, multi-stage process where the model (or human) can request clarifications or obtain contextual hints. The BRAINTEASER(S) SemEval Task, together with its source BRAINTEASER task, is the first step toward injecting AI systems with lateral thinking ability. We hope that the competition results and analysis can inspire future research on developing and evaluating lateral thinking models. 
\section*{Ethical Considerations}
As our brain teasers are ``folk knowledge'' and are published on a range set of websites, it is hard to check their original licenses comprehensively. Yet, the website owners declare permission to print and download material for \textbf{non-commercial use} without modification on the material's copyright. Therefore, we provide the corresponding copyright statements and website URLs for each original brain teaser and its adversarial version. In addition, we ask the task participants to sign a document claiming that the only aim of the data usage is research.
We note that, despite our best efforts, the task data may still contain bias in terms of gender or politics. We will indicate that future research should use the task data with caution.
\section{Acknowledgements} We appreciate Baktash Ansari, Dilip Venkatesh, Soumya Smruti Mishra, Harshit Gupta, and Pouya Sadeghi for their support as emergency reviewers for the competition. This research was sponsored by the Defense Advanced Research Projects Agency via Contract HR00112390061, Defense Advanced Research Projects Agency with award N660011924033 and Strengthening Teamwork for Robust Operations in Novel Groups via number W911NF-19-S-0001.

\bibliography{anthology,custom,2024.semeval2024-1.0}

\newpage

\appendix

\section{CodaLab Leaderboard}
\label{sec:Codalab}
\begin{figure}[h]
	\includegraphics[width=0.48\textwidth]{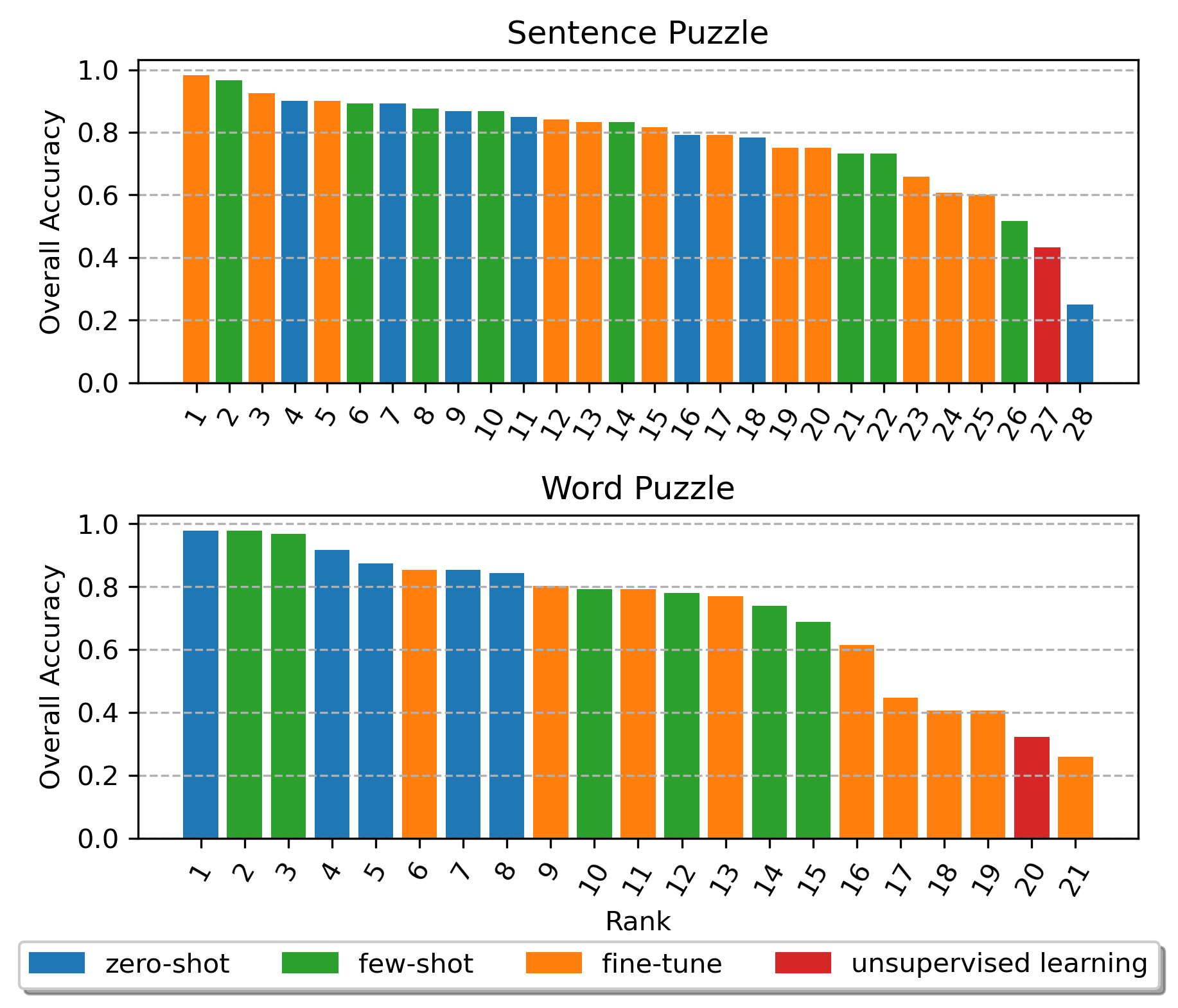}
	\caption{The overall accuracy performance of each team based on method adaptations.} 
	\label{fig:method}
\end{figure}

In the main part of the paper, we only analyse the results for part of the participants' submission due to page limitation. Table \ref{tab:sp_codalab} and \ref{tab:wp_codalab} show a complete set of user names and results of the participants in the CodaLab competition for two subtasks, including users who did not submit a system description. The human evaluation is computed over 102 data randomly sampled from the \textbf{ whole dataset}. The random base is average over three different seeds. The ChatGPT and RoBERTa-L baseline is computed over the whole dataset using OPENAI API\footnote{https://platform.openai.com/docs/api-reference} from 2023/5/01 to 2023/5/15. \\
We visualize each team's overall accuracy in each subtask according to the model adaptation category in Figure~\ref{fig:method}.  In Sentence Puzzle, 12 teams employed fine-tuning, and 15 adopted zero/few-shot approaches. Fine-tuning achieved 1st, 3rd, and 5th positions on the leaderboard, whereas zero/few-shot have 7 places in the top ten. For Word Puzzle, 9 teams used fine-tuning, and 11 opted for zero/few-shot, with the latter dominating the top five ranks, outperforming fine-tuning. 

\begin{table*}[t]
  \setlength\tabcolsep{5pt}
    \caption{Oveview of results of Sentence-puzzle subtask, including user submissions without system description papers. Ori = Original, Sem = Semantic, Con = Context. Team name with (*) submitted the system description paper. The first, second and third submissions per category are represented by \hl{highlight}, \textbf{bold} and  \underline{underline}, respectively.} 
   \label{tab:sp_codalab}
  \begin{center}
   \small
   \begin{tabular}{ l |c | ccc|cc}
   \hline
    \multirow{2}{*}{\textbf{Team Name}} & \multirow{2}{*}{\textbf{Overall}} & \multicolumn{3}{c|}{Instance-based} & \multicolumn{2}{c}{Group-based} \\
     \cline{3-7}
  &  & \textbf{Original}     &    \textbf{Semantic}   &    \textbf{Context}  &  \textbf{Ori \& Sem }  &  \textbf{Ori \& Sem \& Con }           \\              
   \hline
        Abdelhak*& \hl{0.983} & \hl{1.000} & \hl{1.000} & \hl{0.950} & \hl{1.000} & \hl{0.950} \\
        HW-TSC*& \textbf{0.967} & \hl{1.000} & \textbf{0.975} & \textbf{0.925} & \textbf{0.975} & \underline{0.900}  \\
        Maxine& \underline{0.958} & \textbf{0.975} & \textbf{0.975} & \textbf{0.925} & \underline{0.950} & \underline{0.900}  \\
        YingluLi& 0.950 & \textbf{0.975} & \underline{0.950} & \textbf{0.925} & \underline{0.950} & \underline{0.900}  \\
        Theo& 0.950 & \underline{0.950} & \underline{0.950} & \hl{0.950} & \underline{0.950} & \textbf{0.925}  \\
        somethingx95& 0.942 & \underline{0.950} & \underline{0.950} & \textbf{0.925} & \underline{0.950} & \underline{0.900}  \\
        gerald& 0.942 & \underline{0.950} & \underline{0.950} & \textbf{0.925} & \underline{0.950} & \underline{0.900}  \\
        AmazUtah\_NLP*& 0.925 & 0.925 & \underline{0.950} & \underline{0.900} & 0.925 & 0.875  \\
        BITS\,Pilani*& 0.900 & \textbf{0.975} & 0.925 & 0.800 & 0.925 & 0.775  \\
        ALF*& 0.900 & 0.925 & \underline{0.950} & 0.825 & 0.925 & 0.825  \\
        uTeBC-NLP*& 0.892 & \textbf{0.975} & 0.875 & 0.825 & 0.850 & 0.750  \\
        jkarolczak& 0.892 & \textbf{0.975} & 0.875 & 0.825 & 0.875 & 0.775  \\
        kubapok& 0.892 & 0.925 & 0.900 & 0.850 & 0.900 & 0.825  \\
        yangqi*& 0.892 & 0.900 & 0.900 & 0.875 & 0.900 & 0.875  \\
        Mothman*& 0.875 & \textbf{0.975} & 0.850 & 0.800 & 0.850 & 0.700  \\
        zero\_shot\_is\_all\_you\_need*& 0.867 & \underline{0.950} & 0.825 & 0.825 & 0.800 & 0.725  \\
        OUNLP*& 0.867 & \underline{0.950} & 0.875 & 0.775 & 0.850 & 0.725  \\
        justingu& 0.850 & \underline{0.950} & 0.825 & 0.775 & 0.825 & 0.700  \\
        BAMO*& 0.850 & 0.900 & 0.825 & 0.825 & 0.825 & 0.700  \\
        YNU-HPCC*& 0.842 & 0.900 & 0.825 & 0.800 & 0.825 & 0.725  \\
        FtG-CoT*& 0.833 & 0.900 & 0.825 & 0.775 & 0.800 & 0.675  \\
        MasonTigers*& 0.833 & 0.850 & 0.825 & 0.825 & 0.800 & 0.700  \\
        AILS-NTUA*& 0.817 & 0.850 & 0.825 & 0.775 & 0.825 & 0.700  \\
        RiddleMaster*& 0.792 & 0.800 & 0.775 & 0.800 & 0.725 & 0.650  \\
        UMBCLU*& 0.792 & 0.750 & 0.850 & 0.775 & 0.725 & 0.600  \\
        johnp& 0.783 & 0.850 & 0.775 & 0.725 & 0.750 & 0.675  \\
        MABUSETTEH& 0.783 & 0.800 & 0.775 & 0.775 & 0.775 & 0.700  \\
        KnowComp*& 0.783 & 0.825 & 0.775 & 0.750 & 0.725 & 0.625  \\
        ehsan.tavan& 0.775 & 0.800 & 0.800 & 0.725 & 0.775 & 0.675  \\
        amr8ta& 0.775 & 0.775 & 0.775 & 0.775 & 0.750 & 0.650  \\
        yiannispn& 0.767 & 0.800 & 0.800 & 0.700 & 0.750 & 0.625  \\
        haha123& 0.758 & 0.825 & 0.775 & 0.675 & 0.750 & 0.625  \\
        adriti& 0.758 & 0.750 & 0.725 & 0.800 & 0.725 & 0.675  \\
        TienDat23& 0.758 & 0.725 & 0.800 & 0.750 & 0.675 & 0.525  \\
        Deja\_Vu*& 0.750 & 0.775 & 0.700 & 0.775 & 0.700 & 0.625  \\
        NIMZ*& 0.750 & 0.750 & 0.725 & 0.775 & 0.700 & 0.675  \\
        iREL*& 0.733 & 0.775 & 0.725 & 0.700 & 0.700 & 0.575  \\
        GeminiPro*& 0.733 & 0.750 & 0.750 & 0.700 & 0.700 & 0.600  \\
        caoyongwang& 0.725 & 0.800 & 0.700 & 0.675 & 0.700 & 0.550  \\
        IIMAS*& 0.658 & 0.650 & 0.675 & 0.650 & 0.600 & 0.500  \\
        IUST-NLPLAB*& 0.608 & 0.625 & 0.625 & 0.575 & 0.625 & 0.500  \\
        ROSHA*& 0.600 & 0.625 & 0.575 & 0.600 & 0.500 & 0.375  \\
        Team DaVinci*& 0.517 & 0.575 & 0.550 & 0.425 & 0.500 & 0.300  \\
        StFX-NLP*& 0.433 & 0.425 & 0.400 & 0.475 & 0.350 & 0.200  \\
        Team 9& 0.250 & 0.275 & 0.275 & 0.200 & 0.100 & 0.000  \\
        DeBERTa*& 0.250 & 0.225 & 0.250 & 0.275 & 0.200 & 0.075  \\
        amirhallaji& 0.242 & 0.225 & 0.200 & 0.300 & 0.050 & 0.025  \\
        maryam.najafi& 0.233 & 0.225 & 0.275 & 0.200 & 0.100 & 0.025  \\

    \hline
        Human~\cite{jiang-etal-2023-brainteaser}      & 0.920  & 0.907 & 0.907 & 0.944 & 0.907 & 0.889  \\ 
        GPT-4~(BREAINTEASER)     & 0.898 & 0.942 & 0.900 & 0.852 & 0.880 & 0.775   \\ 
        GPT-4~(BREAINTEASER(S))     & 0.858 & 0.925 & 0.825 & 0.825 & 0.8 & 0.775   \\ 
        ChatGPT~(BREAINTEASER)       & 0.627 & 0.608 & 0.593 & 0.679 & 0.507 & 0.397   \\ 
         RoBERTa-L~(BREAINTEASER)   & 0.434 & 0.435 & 0.402 & 0.464 & 0.330 & 0.201 \\
        Random      & 0.244  & 0.255 & 0.249 & 0.228 & 0.056 & 0.014  \\ 
    \hline

    \end{tabular}
  \end{center}
\end{table*}

\begin{table*}[t]
  \setlength\tabcolsep{5pt}
    \caption{Oveview of results of Word-puzzle subtask, including user submissions without system description papers. Ori = Original, Sem = Semantic, Con = Context. Team name with (*) submitted the system description paper. The first, second and third submissions per category are represented by \hl{highlight}, \textbf{bold} and  \underline{underline}, respectively.} 
   \label{tab:wp_codalab}
   \begin{center}
   \small
   \begin{tabular}{ l |c | ccc|cc}
   \hline
    \multirow{2}{*}{\textbf{Team Name}} & \multirow{2}{*}{\textbf{Overall}} & \multicolumn{3}{c|}{Instance-based} & \multicolumn{2}{c}{Group-based} \\
     \cline{3-7}
  &  & \textbf{Original}     &    \textbf{Semantic}   &    \textbf{Context}  &  \textbf{Ori \& Sem }  &  \textbf{Ori \& Sem \& Con }           \\              
   \hline
        Theo& \hl{0.990} & \hl{1.000} &\hl{1.000} & \textbf{0.969} & \hl{1.000} & \hl{0.969} \\
        gerald& \hl{0.990} & \hl{1.000} & \hl{1.000} & \textbf{0.969} & \hl{1.000} & \hl{0.969}   \\
        somethingx95& \textbf{0.979} & \hl{1.000} & \hl{1.000} & \underline{0.938} & \hl{1.000} & \textbf{0.938}  \\
        zero\_shot\_is\_all\_you\_need*& \textbf{0.979} & \hl{1.000} & \hl{1.000} & \underline{0.938} & \hl{1.000} & \textbf{0.938}  \\
        MasonTigers*& \textbf{0.979} & \textbf{0.969} & \textbf{0.969} & \hl{1.000} & \textbf{0.969} & \hl{0.969}   \\
        HW-TSC*& \underline{0.969} & \textbf{0.969} & \underline{0.938} & \hl{1.000} & \underline{0.938} & \textbf{0.938}  \\
        Maxine& \underline{0.969} & \textbf{0.969} &\underline{0.938} & \hl{1.000} & \underline{0.938} & \textbf{0.938}  \\
        YingluLi& \underline{0.969} & \textbf{0.969} & \underline{0.938} & \hl{1.000} & \underline{0.938} & \textbf{0.938}  \\
        kubapok& 0.948 & 0.906 & \hl{1.000} & \underline{0.938} & 0.906 & \underline{0.844}  \\
        BITS Pilani*& 0.917 & \underline{0.938} & \underline{0.938} & 0.875 & \underline{0.938} & 0.812  \\
        justingu& 0.917 & \underline{0.938} & \underline{0.938} & 0.875 & 0.906 & 0.781  \\
        jkarolczak& 0.875 & 0.906 & \underline{0.938} & 0.781 & 0.875 & 0.688  \\
        yangqi*& 0.875 & 0.906 & \underline{0.938} & 0.781 & 0.906 & 0.688  \\
        ehsan.tavan& 0.875 & 0.906 & 0.875 & 0.844 & 0.812 & 0.750  \\
        AILS-NTUA*& 0.854 & 0.875 & 0.906 & 0.781 & 0.812 & 0.719  \\
        johnp& 0.854 & 0.875 & 0.906 & 0.781 & 0.812 & 0.719  \\
        caoyongwang& 0.854 & 0.844 & 0.844 & 0.875 & 0.781 & 0.719  \\
        KnowComp*& 0.854 & 0.844 & 0.906 & 0.812 & 0.844 & 0.656  \\
        RiddleMaster*& 0.844 & 0.844 & 0.844 & 0.844 & 0.781 & 0.656  \\
        yiannispn& 0.833 & 0.844 & 0.844 & 0.812 & 0.719 & 0.625  \\
        AmazUtah\_NLP*& 0.802 & 0.844 & 0.812 & 0.750 & 0.781 & 0.594  \\
        OUNLP*& 0.792 & 0.781 & 0.812 & 0.781 & 0.719 & 0.531  \\
        UMBCLU*& 0.792 & 0.781 & 0.750 & 0.844 & 0.719 & 0.625  \\
        TienDat23& 0.792 & 0.844 & 0.750 & 0.781 & 0.750 & 0.625  \\
        GeminiPro*& 0.781 & 0.781 & 0.719 & 0.844 & 0.594 & 0.594  \\
        YNU-HPCC*& 0.771 & 0.781 & 0.719 & 0.812 & 0.719 & 0.625  \\
        iREL*& 0.740 & 0.719 & 0.719 & 0.781 & 0.562 & 0.531  \\
        Team DaVinci*& 0.688 & 0.719 & 0.719 & 0.625 & 0.594 & 0.469  \\
        Abdelhak*& 0.615 & 0.625 & 0.625 & 0.594 & 0.562 & 0.406  \\
        amr8ta& 0.604 & 0.625 & 0.625 & 0.562 & 0.594 & 0.438  \\
        adriti& 0.604 & 0.656 & 0.625 & 0.531 & 0.625 & 0.375  \\
        MABUSETTEH& 0.583 & 0.594 & 0.625 & 0.531 & 0.562 & 0.281  \\
        NIMZ*& 0.448 & 0.438 & 0.469 & 0.438 & 0.406 & 0.219  \\
        Deja\_Vu*& 0.406 & 0.375 & 0.469 & 0.375 & 0.344 & 0.125  \\
        ROSHA*& 0.406 & 0.438 & 0.375 & 0.406 & 0.375 & 0.250  \\
        StFX-NLP*& 0.323 & 0.406 & 0.219 & 0.344 & 0.125 & 0.062  \\
        IIMAS*& 0.260 & 0.250 & 0.250 & 0.281 & 0.125 & 0.062  \\

    \hline\ 
        Human~\cite{jiang-etal-2023-brainteaser}    & 0.917  & 0.917 & 0.917 & 0.917 & 0.917 & 0.896   \\
        GPT-4~(BREAINTEASER)       & 0.736 & 0.811 & 0.756 & 0.640 & 0.689 & 0.494   \\ 
        GPT-4~(BREAINTEASER(S))       & 0.854 & 0.875 & 0.875 & 0.813 & 0.781 & 0.625   \\ 
        ChatGPT~(BREAINTEASER)        & 0.535 & 0.561 & 0.524  & 0.518 & 0.439 & 0.293  \\ 
        RoBERTa-L~(BREAINTEASER)   & 0.207 & 0.195 & 0.195 & 0.232 & 0.146 & 0.061   \\
        Random       & 0.260 & 0.279 & 0.225 & 0.073 & 0.018 & 0.253  \\ 
    \hline

    \end{tabular}           
   \end{center}
\end{table*}

\section{Participant Systems}
In this section, we list the systems of all participants who submitted a system description paper. The \textbf{team name} represents each system, appended with the corresponding rank in [bracket], keywords in (parentheses), and a short description for further reference. \textit{SP X} and\textit{WP X} represent the ranks in sentence and word puzzles based on overall performance, respectively. 

\textbf{Abdelhak}~[SP\,1;WP\,16]~\cite{kelious-okirim:2024:SemEval2024}~(\textit{Fine-tuned};\textit{DeBERTa};\textit{Zero-shot};\textit{ChatGPT};\textit{Temperature\,Anlysis}) They fine-tuned the pre-trained language model DeBERTa-v3-base in the multiple-choice setting. They further experimented with the relationship between temperature and lateral thinking with ChatGPT in a zero-shot setting.

\textbf{HW-TSC}~[SP\,2;WP\,3]~\cite{li-EtAl:2024:SemEval20245}(\textit{Fine-tuned};\textit{Mixstral};\textit{Zero-shot};\textit{Few-shot};\textit{GPT-3.5};\textit{GPT-4};\textit{Prompting Engineering};\textit{Ensemble}) They first experimented with fine-tuning Mixtral overall whole training set. They turned to GPT-3.5 and GPT-4 due to poor fine-tuning results. They identified and categorized over 20 challenging training instances to include in an extended prompt. Finally, they submitted their result with GPT-4 in the few-shot setting with a well-designed prompting demonstration as well as the ensemble method.

\textbf{AmazUtah\_NLP}~[SP\,6;WP\,10]~\cite{mishra-ghashami:2024:SemEval2024}~(\textit{Fine-tuned};\textit{DeBERTa};\textit{BERT};\textit{External\ Data};\textit{Synthetic\ Data};\textit{RiddleSense}) They fine-tuned DeBERTa and BERT in the multiple-choice setting. They utilized the public puzzle dataset RiddleSense as well as creating humor-style data by prompting GPT 4 as the external dataset. They also experimented by adding commonsense datasets SWAG  and CODAH but found the introduction reduced overall performance.

\textbf{BITS Pilani}~[SP\,7;WP\,5]~\cite{venkatesh-sharma:2024:SemEval2024}~(\textit{Zero-shot};\textit{GPT-4};\textit{Prompting\,Engineering}) They used OpenAI’s GPT-4 model along with prompt engineering in the zero-shot setting to solve these brainteasers.

\textbf{ALF}~[SP 7]~\cite{farokh-zeinali:2024:SemEval2024}~(\textit{Fine-tuned};\textit{ALBERT};\textit{RoBERTa};\textit{DeBERTa};\textit{Flan T5};\textit{Unified QA};\textit{External\ Data};\textit{RiddleSense}) Their experiments focused on two prominent families of pre-trained models, BERT and T5, and fine-tuned ALBERT, RoBERTa, DeBERTa, Flan T5 and Unified QA in the multiple-choice setting. They explored the potential benefits of multi-task finetuning on commonsense reasoning datasets, including RiddleSense, CSQA, PIQA, SIQA, Hellaswag, and SWAG, to enhance performance. 

\textbf{uTeBC-NLP}~[SP\,8]~\cite{sadeghi-abaskohi-yaghoobzadeh:2024:SemEval2024}~(\textit{Fine-tuned};\textit{Zephyr-7B-$\beta$};\textit{Zero-shot};\textit{Few-shot};\textit{GPT-3.5};\textit{GPT-4};\textit{RAG};\textit{External\,Data};\textit{Synthetic\,Data};\textit{Prompting Engineering};\textit{COT};\textit{Lateral thinking enhancement analysis}) They explored Chain of Thought (CoT) strategies, enhancing prompts with detailed task descriptions, and retrieval augmented generation for generating in-context samples. Their experiments involve GPT-3.5 and GPT-4. They also showcased that fine-tuning Zephyr-7B-$\beta$ with a lateral thinking approach significantly enhances the model’s performance on other commonsense datasets.

\textbf{yangqqi}~[SP\,8;WP\,6]~\cite{yang-EtAl:2024:SemEval2024}~(\textit{Zero-shot};\textit{ChatGPT};\textit{RAG};\textit{Self-Adaptive ICL};\textit{Prompting\ Engineering};\textit{External\ Data};\textit{ConceptNet}) They proposed the SHTL system to mimic human lateral thinking ability for solving brain teaser questions. They first retrieved related knowledge concepts from ConceptNet and used SAICL to find the optimal organization for each single test sample. At last, they provide ChatGPT with the related knowledge concepts and find the options to solve the conflicts contained in the related knowledge concepts effectively.

\textbf{Mothman}~[SP\,9]~\cite{chen-groshan-vonbayern:2024:SemEval2024}~(\textit{Zero-shot};\textit{Few-shot};\textit{GPT-4};\textit{Prompting\,Engineering};\textit{COT};) They proposed a system for iterative chain-of-thought prompt engineering which optimizes prompts using a flexible evaluation strategy on both model outputs and input data. They obtain feedback from human evaluation to modify the prompting demonstration interactively to guide GPT-4 to focus on challenging problems. They also proposed a new COT strategy requiring  GPT-4 to produce rationals for both correct and incorrect options.

\textbf{Zero\_Shot\_is\_All\_You\_Need}~[SP\,10;WP\,2]
~\cite{moosavimonazzah-feghhi:2024:SemEval2024}~(\textit{Zero-shot};\textit{Bing};\textit{Gemini};\textit{Mixtral};\textit{Mixtral};\textit{ChatGPT};\textit{Phi-2};\textit{Prompting\,Engineering};\textit{Ensemble};\textit{Debate}) They  examined the zero-shot ability of current state-of-the-art LLMs, Bing, Gemini, Mixtral, ChatGPT and Phi-2 to solve this task. They also tried ensemble and debate prompting engineering methods. 

\textbf{OUNLP}~[SP\,10;WP\,11]~\cite{saravanan-wilson:2024:SemEval2024}~(\textit{Zero-shot};\textit{Few-shot};\textit{GPT-3.5};\textit{GPT-4};\textit{Gemini};\textit{languagemodels};\textit{Prompting Engineering};\textit{COT};\textit{RECONCILE};\textit{External Data};\textit{crawled riddles}) They experimented with a series of structured prompts ranging from basic to those integrating task descriptions and explanations(COT). They use the most similar or the most different training example as the demonstration in the one-shot prompting. They downloaded a collection of riddles from the web as an external data source. In the end, they simulated a council scenario to evoke discussion between different models but didn't observe significant improvement.

\textbf{BAMO}~[SP\,11]~\cite{ansari-rostamkhani-eetemadi:2024:SemEval2024}~(\textit{Fine-tuned};\textit{RoBERTa};\textit{BERT};\textit{Zero-shot};\textit{Open Chat};\textit{Llama-2-70b};\textit{Mixtral};\textit{GPT3.5};\textit{Claud};\textit{Microsoft Copilot};\textit{Prompting Engineering};\textit{ReConcile}) They fine-tuned 2 models, BERT and RoBERTa Large, and employed a Chain of Thought (CoT) zero-shot prompting approach with 6 large language models, such as GPT-3.5, Mixtral, and Llama2. Finally, they utilized ReConcile prompting amount three models.

\textbf{YNU-HPCC}~[SP\,12;WP\,13]~\cite{wang-wang-zhang:2024:SemEval2024}~(\textit{Fine-tuned};\textit{DeBERTa};\textit{External Data};\textit{Back translation}) They fine-tuned DeBERTa in different training strategies and enhanced the training set with back translation. 

\textbf{FtG-CoT}~[SP\,13]~\cite{zhang-ahmed-martin:2024:SemEval2024}~(\textit{Fine-tuned};\textit{BERT};\textit{Zero-shot};\textit{Few-shot};\textit{GPT-3.5};\textit{Prompting Engineering};\textit{COT})  They first fine-tuned BERT in a multi-class classification setting and fine-tuned GPT-3.5 with chain-of-thought generated by zero-shot prompting. Then they picked the set of training demonstrations provided in the few-shot prompt based on the BERT encoding cosine similarity to the test question.

\textbf{MasonTigers}~[SP\,13;WP\,2]~\cite{raihan-EtAl:2024:SemEval2024}~(\textit{Zero-shot};\textit{Few-shot};\textit{GPT-4.5};\textit{Claude};\textit{Mixtral};\textit{Prompting Engineering};\textit{COT})  They explored various prompting strategies to guide the models, including zero-shot, few-shot, and chain-of-thought prompting. The Ensemble method was adopted to enhance COT performance.

\textbf{AILS-NTUA}~[SP\,14;WP\,7]~\cite{panagiotopoulos-EtAl:2024:SemEval2024}~(\textit{Fine-tuned};\textit{DeBERTa};\textit{RoBERTa};\textit{BERT};\textit{Mixtral};\textit{Llama 2};\textit{Phi-2})  They evaluated a plethora of pre-trained transformer-based language models of different sizes and pre-train dataset through fine-tuning. They also delved into models’ frequent failures to obtain a deeper understanding of reasoning cues that make models struggle the most.

\textbf{RiddleMaster}~[SP\,15;WP\,8]~\cite{take-tran:2024:SemEval2024}~(\textit{Fine-tuned};\textit{Mistral};\textit{Zero-shot};\textit{GPT-4};\textit{Prompting Engineering};\textit{COT};\textit{Ensemble}) They compared multiple zero-shot approaches using GPT-4 as well as fine-tuned Mistral output.

\textbf{UMBCLU}\footnote{The paper was withdrawn.}~[SP\,15;WP\,11]~(\textit{Fine-tuned};\textit{Flan-T5};\textit{Data Augmentation}) They fine-tuned and evaluated various T5 family models on both the word and sentence puzzle tasks and showed that training on the alternative contexts improves a model’s lateral reasoning capability.

\textbf{KnowComp}~[SP\,16;WP\,7]~\cite{wang-EtAl:2024:SemEval20243}~(\textit{Zero-shot};\textit{ChatGPT};\textit{Prompting Engineering}) They first prompted ChatGPT to identify relevant instances in the question and generate conceptualizations for the identified instances. They then converted each puzzle into a declarative format and modified the task to involve selecting the most plausible statement from the options.

\textbf{NIMZ}~[SP\,20;WP\,19]~\cite{rahimi-EtAl:2024:SemEval20242}~(\textit{Fine-tuned};\textit{BERT};\textit{RoBERTa};\textit{T5};\textit{QA-GNN};\textit{External Data};\textit{ConceptNet}) They fine-tuned BERT, RoBERTa and T5 and evaluated their performance. They used ConceptNet as an external knowledge source and fine-tuned graph neural network QA-GNN and suggested its superiority on sentence puzzle.

\textbf{Deja-Vu}~[SP\,20;WP\,20]~\cite{chakraborty-rahman-faruqe:2024:SemEval2024}~(\textit{Fine-tuned};\textit{BERT};\textit{RoBERTa};\textit{XLNet};\textit{BART};\textit{T5};\textit{Data Augmentation}) They fine-tuned five transformer-based language models and found the integration of sentence and word puzzles into a single dataset led to a noticeable decrease in accuracy.

\textbf{GeminiPro}~[SP\,21;WP\,12]~\cite{choi-na:2024:SemEval2024}~(\textit{Zero-shot};\textit{Few-shot};\textit{Gemini};\textit{Prompting Engineering}) They tested Gemini's performance in zero-shot and few-shot settings. They experimented with whether tailor-made demonstrations to specific tasks can alleviate confusion and aid in 049 problem-solving.

\textbf{iREL}~[SP\,21;WP\,14]~\cite{gupta-EtAl:2024:SemEval2024}~(\textit{Zero-shot};\textit{Few-shot};\textit{Gemini};\textit{Prompting Engineering};\textit{COT}) They tested Gemini's performance in zero-shot and few-shot settings. Especially in the few-shot setting, reasoning from Gemini and GPT-4 are integrated into the demonstration, selected by static or dynamic strategy.

\textbf{IIMAS}~[SP\,23;WP\,22]~\cite{reyes-ramosflores-martnezmaqueda:2024:SemEval2024}~(\textit{Fine-tuned};\textit{BERT};\textit{RoBERTa};\textit{ChatGPT};\textit{Gemini};\textit{Data Augmentation}) They tackled this challenge by applying fine-tuning techniques with pre-trained models (BERT and RoBERTa Winogrande) while also augmenting the dataset with the LLMs ChatGPT and Gemini. During the training, they transformed the data format for specific templates.

\textbf{IUST-NLPLAB}~[SP\,24]~\cite{abbaspour-moosavimonazzah-eetemadi:2024:SemEval2024}~(\textit{Fine-tuned};\textit{MPNET};\textit{Zero-shot};\textit{GPT-3.5}) They first introduced a zero-shot approach leveraging the capabilities of the GPT3.5 model. Additionally, they presented three finetuning methodologies utilizing MPNET as the underlying architecture, each employing a different loss function.

\textbf{ROSHA}~[SP\,25;WP\,20]~\cite{rostamkhani-mousavinia-eetemadi:2024:SemEval2024}~(\textit{Fine-tuned};\textit{RoBERTa};\textit{Zero-shot};\textit{GPT-3.5};\textit{Gemini};\textit{Mixtral};\textit{GPT-4};\textit{External Data};\textit{BiRdQA};\textit{RiddleSense};\textit{Prompting Engineering};\textit{Reconcile}) They applied the XLM-RoBERTa model both to the original training dataset and concurrently to the original dataset alongside the BiRdQA dataset and the RiddleSense for comprehensive model training. They also tested the Reconcile prompting strategy with GPT-3.5, Gemini as well as Mixtral and zero-shot on GPT-4.

\textbf{DaVinci}~[SP\,26;WP\,15]~\cite{mathur-jindal-shrivastava:2024:SemEval2024}~(\textit{Few-shot};\textit{GPT-3.5};\textit{Prompting Engineering}) They used few-shot prompting on GPT-3.5 with rationale and gained insights regarding the difference in the nature of the two types of questions.

\textbf{StFX-NLP}~[SP\,27;WP\,21]~\cite{heavey-hughes-king:2024:SemEval2024}~(\textit{unsupervised};\textit{External Data};\textit{WordNet}) They explored three unsupervised learning models. Two of these models incorporate word sense disambiguation and part-of-speech tagging, specifically leveraging SensEmBERT and the Stanford log-linear part-of-speech tagger. The third model relies on a more traditional language modelling approach.


\textbf{DeBERTa}~[SP\,28]~\cite{siino:2024:SemEval20246}~(\textit{Zero-shot};\textit{DeBERTa}) They used DeBERTa in zero-shot setting.

\end{document}